\documentclass{article}

\usepackage{arxiv}

\usepackage[utf8]{inputenc} % allow utf-8 input
\usepackage[T1]{fontenc}    % use 8-bit T1 fonts
\usepackage{hyperref}       % hyperlinks
\usepackage{url}            % simple URL typesetting
\usepackage{booktabs}       % professional-quality tables
\usepackage{amsfonts}       % blackboard math symbols
\usepackage{nicefrac}       % compact symbols for 1/2, etc.
\usepackage{microtype}      % microtypography
\usepackage{lipsum}		% Can be removed after putting your text content
\usepackage{graphicx}
\usepackage{natbib}
\usepackage{doi}

\title{Improving Ethical Outcomes with Machine-in-the-Loop: Broadening Human Understanding of Data Annotations\thanks{{This paper is presented at the Human Centered AI workshop at the 35th Conference on Neural Information Processing Systems (NeurIPS), Dec 13th 2021.}}}

\author{ \href{https://orcid.org/0000-0002-7446-4639}{\includegraphics[scale=0.06]{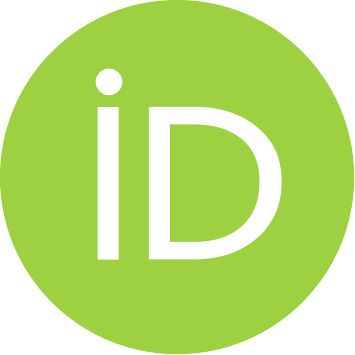}\hspace{1mm}Ashis Kumer Biswas} \\
	Computer Science \& Engineering,\\
	University of Colorado Denver,\\
	Colorado, CO 80234, \\
	\texttt{ashis.biswas@ucdenver.edu} \\
	\And
	Geeta Verma\\
	School of Education and Human Development,\\
    University of Colorado Denver\\
    Denver, CO 80204 \\
    \texttt{geeta.verma@ucdenver.edu} \\
	\AND
	Justin Otto Barber \\
    Radiology Partners\\
    El Segundo, CA 90245\\
    \texttt{justin.barber@radpartners.com}
}

\date{December 13, 2021}

\renewcommand{\shorttitle}{Improving Ethical Outcomes With Machine-in-the-Loop}

\hypersetup{
pdftitle={Machine-in-the-loop to improve ethical outcomes},
pdfsubject={Biases, ethics, AI},
pdfauthor={Ashis Kumer Biswas, Geeta Verma, Justin Otto Barber},
pdfkeywords={AI, Biases, Ethics, Machine-in-the-loop},
}

\begin{document}
\maketitle

\begin{abstract}
We introduce a machine-in-the-loop pipeline that aims to address root causes of unwanted bias in natural language based supervised machine learning tasks in the education domain. Learning from the experiences of students is foundational for education researchers, and academic administrators. 21st-century skills learned from experience are becoming a core part of college and career readiness as well as the hiring process in the new knowledge economy. Minoritized students demonstrate these skills in their daily lives, but documenting, assessing, and validating these skills is a huge problem for educational institutions. As an equity focused online platform, LivedX translates minoritized students’ lived experiences into the 21st century skills, issues micro-credentials, and creates personal 21st century skills portfolio. To automate the micro credential mining from the natural language texts received from the students’ submitted essays, we employed a bag-of-word model to construct a multi-output classifier. Despite our goal, our model initially exacerbated disparate impact on minoritized students. We used a machine-in-the-loop model development pipeline to address the problem and refine the aforementioned model to ensure fairness in its prediction.
\end{abstract}

% keywords can be removed

\section{Introduction \label{sec:introduction}}

\textbf{The problem of data annotations from academic environments}: 
As the result of a long and complex history of marginalization (e.g., racism, classism, oppression), minoritized students consistently lag behind their majority peers in academic achievement and career readiness. This lag has been influenced by structural inequalities such as those framed by poverty, migration, and/or undocumented status or homelessness \citep{massey1993american}. This lag has also been called an opportunity gap because, in part, minoritized students lack access to the majority peers’ experiences and the discourses upon which educational achievement is defined \citep{carter2013closing}. As such, the lived experiences of minoritized students are often defined in terms of deficits and do not carry the cultural capital found in more privileged students’ experiences like attending AP courses, or early college, received funded extracurricular experiences. This marginalization leads to problems when natural language datasets for supervised learning are annotated. Minority perspectives in data samples may be defined in terms of what the minority speaker lacks rather than what they bring to the table. Many minoritized students bring a unique set of life experiences (e.g., resilience, perseverance, stress management, and conflict resolution) that have provided them with skills and competencies to provide alternative affordances in the area of 21st-century skills (e.g., bilingualism, problem-solving, critical thinking, collaboration). We propose an data annotation pipeline that has led to better outcomes for minority students in our project. 

\textbf{Improving ethical outcomes for our LivedX online platform: From deficit to asset mindset}:
 In the Pilot phase of this work, the equity focused LivedX online platform was created, and is accessible at \url{https://livedx.com/}. In addition, the core ideas were pilot tested by guiding students to document their lived experiences. A robust LivedX research framework was developed--using a phenomenological perspective \citep{heidegger1962being} and a Funds of Knowledge framework \citep{moll1992funds}--and used to interpret students’ experiences. Micro-credentials were issued to create a portfolio of soft/power skills . This iterative process helped revise and update the LivedX framework. As of now, a total of 170  micro-credentials can be awarded using the LivedX framework.

The purpose of the LivedX framework is to improve outcomes for minoritized students, but we found unintentional and unwanted bias within the framework. We improved our model and the outcomes for minoritized students by creating a dialog between our annotators and our model. Our annotators shape the data used by the model, and our model provides valuable feedback to our human annotators. This feedback provides annotators the opportunity to reconsider their own inherent biases and to account for the way the models leverage their annotations. The only way to achieve ethical results in use cases such as these is to view the entire process iteratively--from end to end. The annotators do not simply decide on annotations once and for all. Rather, they are privy to the sorts of mistakes the machine learning model makes from their annotations in order to consider possible biases in their own annotation work. They consider how to address the root (rather than superficial) of the unwanted bias \citep{DBLP:journals/corr/abs-1808-00023,10.1145/3375627.3375828,DBLP:journals/corr/abs-2011-02279}. The alternative is to ensure that the unfairness generated by an algorithm on particular annotations is distributed equally among all groups. This does not usually improve the situation much because it fails to account for and address the characteristics of the data that are leading to the unwanted bias in the first place. Further, such superficial alternative approaches fail to provide guidance on how the machine learning fueled platform like LivedX should address the root cause of the unwanted biases \citep{10.1145/3375627.3375828}. In this paper, we examine the machine-in-the-loop strategy to understand the intersection of machine and human bias for an effective and fair micro-credentialing process.

\section{Materials and Methods}
\subsection{Dataset}
We collected data through our LivedX interface. Users responded to four prompts with which they can narrate a small experience that took place in their life. The focus of this experience is asset based (e.g., helping others, conflict resolution, taking care of physical health) as we wanted to capture users' positive experiences and reward them for participating in these experiences. As of 2021-08-10, we have collected 2,974 experience essays from 621 student users from our LivedX platform. The length of each natural language text in the collected experience corpus ranges from 9 to 2,649 characters.

The LivedX framework is built to assign from a pool of 152 \textit{a priori} microcredentials that we refer to as ``level-3 codes''. The codes are then grouped into 39 disjoint code groups, we call ``level-2 codes'' in an agglomerative fashion. We further introduced 8 disjoint code groups, we named ``level-1 codings'' from the level-2 codes naturally formed groups of eight which we call the ``level-1 codes''. Each submitted experience usually gets 1--5 level-3 codes based on the robustness of the submitted experience. These codes could be in various categories such as global citizenship, social emotional learning, collaboration, and communication. 

For the purpose of interrater reliability (IRR), we had a group of 3 coders (i.e., human annotators) who annotated each of the submitted experience texts to receive level-3 microcredentials. All three coders had different racial and ethnic backgrounds. Two coders could be classified as people of color and the third coder is identified as Caucasian. In addition, all three coders had diverse professional backgrounds. One coder is trained in Science, Technology, Engineering, and Mathematics (STEM) education and the other two in research methodology. Diverse professional background also brought richness to the coding process.% -- special education, STEM education, and museum education/community outreach coordinator. 
The coders discussed each experience and shared their interpretation of the experience and then the coding team came to a consensus by creating shared meaning ascribed to particular text.

%In this paper, we discuss ways to prepare data for machine learning algorithms by 1)  creating shared meaning-making; 2)recruiting coders with diverse personal backgrounds (persons of color & special education); and 3) recruiting coders with diverse professional backgrounds (informal education, higher education, and k-12 education). We did that intentionally to bring different perspectives to machine learning data preparation and have the machine learning algorithm reflect our value systems and create ____ a positive bias? 

\subsection{Predicting microcredentials}
We adapted a bag-of-word model approach to solve the multi-label text classification problem of predicting a set of appropriate level-3 codes out of 152 codes for a given experience text. To obtain such a model, we first removed common English stop words. We converted the text dataset into the term frequency inverse document frequency (TF-IDF) matrix \citep{ramos2003using}. For simplicity, we chose to build a multi-output classifier. This strategy consists of fitting one binary classifier, e.g., logistic regression, per target; and we have 152 level-3 targets. We conducted a 10-fold cross validation on a split training set to decide on the hyper-parameters of the model.
\subsection{Evaluation with fairness metrics}
We evaluated the microcredential predictions using conditional statistical parity (CSP). CSP measures whether particular groups have equal probability of receiving a favorable outcome (in this case a credential) while permitting a legitimate factor to affect the prediction fairly \citep{DBLP:journals/corr/abs-2102-08453}. In our case, we consider students who have submitted fewer than 5 experiences to be too inexperienced and exclude them from consideration. For level-1 and level-2 credentials, we see CSP for many--but not all--credentials.

We found that CSP exists for most credentials between most groups, but occasionally a group achieves substantially fewer credentials than other groups. For example, the students who identified as white achieved a CSP of 0.1049 for the "working with others" credential, whereas students who identify as black or African American achieved a CSP of 0.0368 (see \ref{tab:csp}).
If we take the data annotations to be authoritative, we might suspect that students who identify as black or African American are somehow deficient (perhaps due to systemic discouragement of these experiences). A second possibility also exists, however. Students who identify as black or African American may express how they work with others in ways that are different from white students and in ways that the annotators might have missed. These questions cannot be answered by the outputs of the model alone. They need to be answered by the subject experts (in this case, the annotators).

For our problem, we mark differences in CSP of greater than 0.05 as credentials of concern.

\subsection{Machine-in-the-loop Pipeline} 

\begin{figure}[!t]
    \centering
    \includegraphics[scale=0.5]{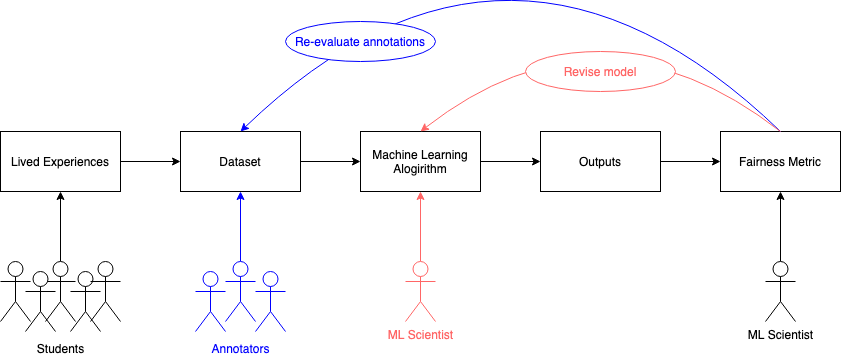}
    \caption{Our proposed machine-in-the-loop fairness pipeline emphasizes two forms of revisions based upon fairness outcomes: (1) a re-evaluation of annotations based upon predictive outcomes and (2) model revision. Note that the Dataset and Machine Learning Algorithm nodes are nodes where human and machine collaborate iteratively.}
    \label{fig:galaxy}
\end{figure}

Figure \ref{fig:galaxy} illustrates our proposed iterative pipeline to refine the machine learning models by sending the intermediate predictions by the model back to the human annotators considering fairness outcomes. Many approaches to enforce fairness across machine learning model prediction emphasize strategies related to algorithm manipulation or dataset balancing to address discrepancies of fairness \citep{10.1145/3457607}. We propose that data annotation should also be an iterative process included in the machine learning pipeline, since annotations are frequently problematic insofar as they can encode societal norms \citep{10.1145/3461702.3462594}. Annotations should periodically be reconsidered in light of fairness metric outcomes, especially for high bias models. High bias models can identify trends in the annotations that result from the annotators' unwanted bias. The predictions of these models can identify trends that need to be explored by the annotators.

\section{Results and Discussion}
\subsection{Microcredential Assignment Experiments}
We trained three models for predicting level-3 microcredentials: Logistic Regression, Nai\"ve Bayes and Support Vector Machines with linear kernel on a 5000 dimensional ``term frequency inverse document frequency'' scores of the experience text corpus that we retrieved from the LivedX platform. We conducted a 10-fold cross-validation for each of the models, and the evaluation results are presented in Table \ref{tab:model-comparison}. Each of the three classifiers show promising prediction performance where Logistic Regression based multi-output classification demonstrated slightly faster performance than the rest on an average. Currently, the model is being used at the LivedX platform to automate the microcredential assignment problem to assist the human annotators.
\begin{table}[!tph]
\begin{center}
\caption{Experimental evaluation of three Bag-of-Word Models in terms of average accuracy of each of predictions of 152 level-3 microcredential classes, and CPU time required for microcredential assignment of each experience text submitted. \label{tab:model-comparison}}
\begin{tabular}{ccc}
\toprule
\textbf{Model name} & \textbf{Average accuracy} & \textbf{Average prediction time per essay (s)}\\
\midrule
Logistic Regression & $0.96\pm 0.066$& 0.089959\\
Support Vector Machines & $0.96\pm0.006$ & 0.093059\\
Nai\"ve Bayes & $0.94 \pm 0.005$
& 0.109604\\\bottomrule
\end{tabular}
\end{center}
\end{table}
\vspace{-0.15in}
\subsection{Evaluation based on Fairness Metrics and Model Refinement}
When we isolate the possible race categories to "white" and "black or African American" for the sake of illustration, 12.8\% of the level-2 microcredentials have a difference in CSP greater than or equal to 0.05. 60\% of these differences favored white submissions. The annotators assessed these differences looking especially for initial annotations that might contain some annotator bias. For instance, Table \ref{tab:csp} outlines effectiveness of model refinement through the proposed iterative pipeline on a particular level-2 microcredential, ``Working with Others''.

After one iteration of training our model, we sent the outcomes back to annotators and asked them to reconsider the annotations for credentials exhibiting a difference in CSP greater than or equal to 0.05 between two demographic groups. This reconsideration was performed one demographic group at a time, so that the annotators could consider possible systemic bias as explanations. For example, the annotators discovered that black or African American students in the sample tend to talk about working with others passively, whereas white students tend to describe working with others actively.
\begin{table}[!tph]
\begin{center}
\caption{Conditional statistical parity for the "Working with Others" credential by race, before and after annotation re-assessment.\ \label{tab:csp}}
\begin{tabular}{ccc}
\toprule
\textbf{Iteration} & \textbf{White} & \textbf{Black or African American}\\
\midrule
Before Annotation Reassessment & $0.1049$& $0.0368$ \\
After Annotation Reassessment & $0.1049$ & $0.0743$ \\\bottomrule
\end{tabular}
\end{center}
\end{table}
\vspace{-0.25in}
%\subsection{Model Refinements}

\section{Conclusion and Future Work}

In this project, we shared an equity serving online platform (LivedX) that translates peoples’ everyday life experiences into micro credentials, which utilizes the machine learning pipeline proposed here. We found that iterative data annotation--with deliberate reflection on the fairness of outcomes for each subgroup--leads to more just outcomes. The project is still in its early stages in terms of scale of the data set and the diversity of population included in the experience essay corpus and the human annotators. The model could also be improved through using sequential neural network models, like BERT instead of bag of words. We hope, through this presentation, that we create a dialog in the community that would help us reflect on best practices for iterative annotations of data that leads to more equitable outcomes for marginalized groups.

\bibliographystyle{unsrtnat}
\bibliography{references}

\end{document}